# Control of a Lightweight Flexible Robotic Arm Using Sliding Modes

**Victor Etxebarria, Arantza Sanz & Ibone Lizarraga**
Dpto. de Electricidad y Electrónica, Facultad de Ciencia y Tecnología,
Universidad del País Vasco, Apdo. 644, 48080 Bilbao, Spain
victor@we.lc.ehu.es

*Abstract: This paper presents a robust control scheme for flexible link robotic manipulators, which is based on considering the flexible mechanical structure as a system with slow (rigid) and fast (flexible) modes that can be controlled separately. The rigid dynamics is controlled by means of a robust sliding-mode approach with well-established stability properties while an LQR optimal design is adopted for the flexible dynamics. Experimental results show that this composite approach achieves good closed loop tracking properties both for the rigid and the flexible dynamics.*

## 1. Introduction

Flexible-link robotic manipulators have many advantages with respect to conventional rigid robots. These mechanisms are built using lighter, cheaper materials, which improve the payload to arm weight ratio, thus resulting in an increase of the speed with lower energy consumption. Moreover these lightweight arms are more safely operated due to the reduced inertia and compliant structure, which is very convenient for delicate assembly tasks and interaction with fragile objects, including human beings.

However, the dynamic analysis and control of flexible-link manipulators is much more complex than the analysis and control of the equivalent rigid manipulators. From the modelling standpoint, the challenges are associated with the fact that the non-linear rigid body motions are now strongly coupled with the distributed effects of the flexibility along the mechanical structure. This coupling varies with the system configuration and the load inertia. Besides, the dynamic equations of flexible structures are infinite dimensional, although for control purposes approximated finite order models are usually considered. This truncation, along with the difficulties in modelling the coupling and nonlinearities of the system, can be the source of uncertainties in the dynamical models, which in turn can lead to poor or unstable control performance.

In addition, for a flexible-link manipulator the number of controlled variables is strictly less than the number of mechanical degrees of freedom, so the control aim is double in this case. First, a flexible manipulator controller must achieve the same motion objectives as a rigid manipulator. Second, it must also stabilize the vibrations that are naturally excited. Moreover, if the tip position is chosen as the system output, flexible links arms are non-minimum phase systems (Wang, D. & Vidyasagar, M. 1991, Moallem, M; Patel, R.V. & Khorasani, K. 2001), which implies that the conventional robot control methods based on feedback linearization cannot be applied to flexible manipulators.

The existence of the slow (rigid) and fast (flexible) modes allows the application of the singular perturbation theory to flexible arms (Kokotovic, P; Khalil, H.K. & O'Reilly, J. 1999). Using this approach, the system dynamics is decomposed into a slow and a fast subsystem. Then, a composite control strategy can be adopted, with the controller having slow and fast terms. Since the slow subsystem has the same structure and properties than the equivalent rigid arm, the slow control can be based on well-established control schemes for rigid manipulators, while the fast control can be synthesized as a linear control with the slow state variables acting as parameters. This combined slow-fast strategy has proved to be a promising control method for robotic applications (Li, Y.; Tang, B.; Zhi, Z. & Lu, Y. 2000, Lizarraga, I. & Etxebarria, V. 2003).

In this paper robust control strategies are applied to a flexi-ble-link manipulator within the singular perturbation framework, in contrast with other approaches that have appeared in the literature, which mainly use adaptive control (Yang, J.H.; Lian, F.L. & Fu, L.C. 1997, Bai, M; Zhou, D.H. & Schwarz, H 1998). In the present paper, a sliding-mode controller is designed for the slow subsystem and an optimal LQR strategy is proposed for the fast subsystem. This kind of sliding control scheme was introduced in (Ba-rambones, O. & Etxebarria, V. 2001) for the tracking control of rigid manipulators. It consists of an adaptive term, which is a feedback linearization law for the modeled dynamics, and a sliding term, which is used to overcome the uncertainties. In order to prevent the high frequency chattering, a smooth function is considered instead of the sign function. An adaptive sliding gain can be used to avoid the necessity of *a priori* knowledge of the unmodeled dynamics and noise bounds. On the other hand, the LQR controller for the fast subsystem is designed to stabilize the deflections of the flexible modes.

The paper is organized as follows. Section 2 gives a finite dimensional model for a generic flexible-link manipulator, suitable for control purposes, together with its approximate representation through the slow and fast subsystems derived from the singular perturbation method. Then, in section 3 the proposed combined sliding-LQR control strategy based on the separate slow-fast representation is presented. Next, the performance of the proposed design is illustrated in section 4 by means of experiments on a laboratory flexible arm. Finally, conclusions are given in the last section.

## 2. Model for flexible-link manipulators

Conventional rigid-link manipulators are modelled as a ser of nonlinear coupled ordinary differential equations (ODEs). However, in the case of flexible manipulators this rigid dynamics is coupled with the distributed effects of the flexibility along the mechanical structure, which lead to a model expressed in partial differential equations (PDEs), where both time and spatial derivatives are relevant. PDEs are not very convenient as models for control design purposes, since they are theoretically equivalent to infinite-dimensional systems. In order to derive a finite-dimensional ODE, the deformation of each link is expressed as a superposition of modes where the spatial and time variables are separated:

$$w_j(x_j,t) = \sum_{i=1}^{m_j} \phi_{ij}(x_j) q_{f_{ij}}(t) \quad i...n \qquad (1)$$

where $q_{f_{ij}}(t)$ is the mode amplitude and $\phi_{ij}(x_j)$ is the mode shape. Natural modes can be analytically calculated for single flexible links (see for instance Fraser, A.R. & Daniel, R.W. 1991). However, in the general multilink case, a set of assumed mode shapes, that satisfy certain geometric boundary conditions, are usually considered. By applying Lagrange formulation and using expression (1) the dynamics of any multilink flexible-link robot can be represented by:

$$M(q)\ddot{q} + V(q,\dot{q}) + Kq + F(q,\dot{q}) + G(q) = B(q)\tau \qquad (2)$$

with $q(t) = \left[q_r^T, q_f^T\right]^T$, where $q_r$ is the vector of rigid modes (generalized joint coordinates) and $q_f$ is the vector of flexible modes, respectively defined as:

$$q_r = \left[q_{r_1},\ldots,q_{r_n}\right]^T$$
$$q_f = \left[q_{f_{11}},\ldots,q_{f_{1m_1}},\ldots,q_{f_{n1}},\ldots,q_{f_{nm_n}}\right]^T \qquad (3)$$

where $n$ is the number of links and $m_i$ are the number of flexible modes considered for each link. $M(q)$ represents the inertia matrix, $V(q,\dot{q})$ is the Coriolis and centrifugal vector, $K$ is the stiffness matrix, $F(q,\dot{q})$ is the friction vector, $G(q)$ is the gravity vector. $B(q)$ is the input matrix, which depends on the particular boundary conditions corresponding to the assumed modes. Finally, $\tau$ includes the control torques applied to each joint.

The dynamical equation (2) can be partitioned according to the rigid and flexible modes:

$$\begin{bmatrix} M_{rr} & M_{rf} \\ M_{fr} & M_{ff} \end{bmatrix} \begin{bmatrix} \ddot{q}_r \\ \ddot{q}_f \end{bmatrix} + \begin{bmatrix} V_{rr} & V_{rf} \\ V_{fr} & V_{ff} \end{bmatrix} \begin{bmatrix} \dot{q}_r \\ \dot{q}_f \end{bmatrix} +$$
$$\begin{bmatrix} 0 & 0 \\ 0 & K_{ff} \end{bmatrix} \begin{bmatrix} q_r \\ q_f \end{bmatrix} + \begin{bmatrix} F_r \\ 0 \end{bmatrix} + \begin{bmatrix} G_r \\ G_f \end{bmatrix} = \begin{bmatrix} B_r \\ B_f \end{bmatrix} \tau \qquad (4)$$

where the following properties are known to be verified, by the Lagrangian structure of the system:

**Property 1:** $M(q)$ and $M_{rr}$ are non-singular, symmetric, positive definite matrices.

**Property 2:** The Coriolis and centrifugal vector $V(q,\dot{q})$ has been expressed as the product of a matrix by a vector:

$$V(q,\dot{q}) = V_m(q,\dot{q})\dot{q} = \begin{bmatrix} V_{rr} & V_{rf} \\ V_{fr} & V_{ff} \end{bmatrix} \begin{bmatrix} \dot{q}_r \\ \dot{q}_f \end{bmatrix} \qquad (5)$$

where $V_m(q,\dot{q})$ and $V_{rr}$ verify that $N = \dot{M} - 2V_m$ and $N_{rr} = \dot{M}_{rr} - 2V_{rr}$ are both antisymmetric matrices.

**Property 3:** If the rigid coordinates $q_r$ are chosen to be the joint angles corresponding to each link (the simpler and most usual choice) then $B_r$ is the identity matrix and $B_f = 0$.

Applying the singular perturbation theory (Kokotovic, P; Khalil, H.K. & O'Reilly, J. 1999) the system dynamics can be decomposed into a slow and a fast subsystem. In particular, a singularly perturbed model of (4) can be obtained by introducing a small-scale factor $\varepsilon$ defined as $\varepsilon^2 = k_m^{-1}$, where $k_m^{-1}$ is the smallest stiffness constant. Also, new fast variables $\psi$ and a scaled stiffness matrix $\tilde{K}_{ff}$ are defined as:

$$\psi = k_m q_f = \frac{1}{\varepsilon^2} q_f \quad K_{ff} = k_m \tilde{K}_{ff} = \frac{1}{\varepsilon^2} \tilde{K}_{ff} \qquad (6)$$

Using these new variables, equation (4) can be rewritten:

$$\begin{aligned}\ddot{q}_r &= B_r^1 \tau - V_{rr}^1 \dot{q}_r - V_{rf}^1 \varepsilon^2 \dot{\psi} - H_{rf} \tilde{K}_{ff} \psi - \\ &\quad - H_{rr} F_r - H_{rr} G_r - H_{rf} G_f \\ \varepsilon^2 \ddot{\psi} &= B_f^1 \tau - V_{fr}^1 \dot{q}_r - V_{ff}^1 \varepsilon^2 \dot{\psi} - H_{ff} \tilde{K}_{ff} \psi - \\ &\quad - H_{fr} F_r - H_{fr} G_r - H_{ff} G_f \end{aligned} \qquad (7)$$

where the new matrices appearing in the expression are defined directly from the old matrices as follows:

$$H(q) = M^{-1}(q); \quad H(q) = \begin{bmatrix} H_{rr} & H_{rf} \\ H_{fr} & H_{ff} \end{bmatrix} \qquad (8)$$

and:

$$\begin{aligned}&B_r^1 = H_{rr} B_r + H_{rf} B_f; \quad B_f^1 = H_{fr} B_r + H_{ff} B_f; \\ &V_{rr}^1 = H_{rr} V_{rr} + H_{rf} V_{fr}; \quad V_{fr}^1 = H_{fr} V_{rr} + H_{ff} V_{fr}; \\ &V_{rf}^1 = H_{rr} V_{rf} + H_{rf} V_{ff}; \quad V_{ff}^1 = H_{fr} V_{rf} + H_{ff} V_{ff}; \end{aligned} \qquad (9)$$

As mentioned in the introduction, a double control objective needs to be achieved. On the one hand, $q_r$ must track a prescribed trajectory. On the other hand, the elastic vibrations of the flexible modes have to be damped out. For that, a composite control law is defined:

$$\tau = \bar{\tau} + \tilde{\tau} \qquad (10)$$

where $\bar{\tau}$ is the slow component and $\tilde{\tau}$ is the fast component. In the following the bar over the variables is used to denote the slow part of them.
The slow subsystem is obtained by setting $\varepsilon = 0$ in (7). The second equation gives the slow manifold equation:

$$\bar{\psi} = \bar{K}_{ff}^{-1} \bar{H}_{ff}^{-1} (\bar{B}_f^1 \bar{\tau} - \bar{V}_{fr} \dot{\bar{q}}_r - \bar{H}_{fr} \bar{F}_r - \\ - \bar{H}_{fr} \bar{G}_r - \bar{H}_{ff} \bar{G}_f) \qquad (11)$$

and the first one gives the slow dynamics, that using expression (11) of the slow manifold it can be written:

$$\ddot{\bar{q}}_r = \bar{M}_{rr}^{-1} \left[ \bar{B}_r \bar{\tau} - \bar{V}_{rr} \dot{\bar{q}}_r - \bar{F}_r - \bar{G}_r \right] \qquad (12)$$

To derive the fast subsystem a time-scale change $T = t/\varepsilon$ is performed and new fast variables are defined:

$$\varphi 1 = \psi - \bar{\psi} \quad \varphi 2 = \varepsilon \dot{\psi} \qquad (13)$$

In this new time scale the slow variables are treated as constants. Thus, from (7) and using (11), the following equations, which define the dynamics of the fast modes near the slow manifold, can be obtained:

$$\left.\begin{aligned}\frac{d\varphi 1}{dT} &= \varphi 2 \\ \frac{d\varphi 2}{dT} &= -\bar{H}_{ff} \bar{K}_{ff} \varphi 1 - \bar{V}_{ff}^1 \varepsilon \varphi 2 - \bar{B}_f^1 \tilde{\tau}\end{aligned}\right\} \qquad (14)$$

Note that the term $\bar{V}_{ff}^1 \varepsilon \varphi 2$, which is first order in the perturbation parameter and is usually neglected, has been explicitly considered in order to take into account the damping of the flexible modes. Now, if two matrices $A_F$ and $B_F$ are defined:

$$A_F = \begin{bmatrix} 0 & I \\ -\bar{H}_{ff} \bar{K}_{ff} & -\bar{V}_{ff}^1 \varepsilon \end{bmatrix}; \quad B_F = \begin{bmatrix} 0 \\ \bar{B}_f^1 \end{bmatrix} \qquad (15)$$

and a fast state vector is defined as $\varphi = [\varphi 1 \quad \varphi 2]^T$, then the fast subsystem can be written as:

$$\frac{d\varphi}{dT} = A_F \varphi + B_F \tilde{\tau} \qquad (16)$$

which describes the fast variables evolution around the slow manifold (11).

## 3. Control design

As pointed out before, the design of a feedback controller for the singularly perturbed model is developed according to a composite control strategy $\tau = \bar{\tau}(\bar{q}_r, \dot{\bar{q}}_r) + \tilde{\tau}(\varphi 1, \varphi 2)$, where $\bar{\tau}$ is the control law for the slow subsystem (12) and $\tilde{\tau}$ is the fast control law that must satisfy the constraint $\tilde{\tau}(0,0) = 0$ (i.e. the fast controller is inactive along the equilibrium manifold (11) ). The design process for the two controllers is described in the following

### 3.1 Sliding-mode control for the slow system

The objective of the slow control consists of ensuring that the slow variables follow a prescribed trajectory. Since the slow subsystem (12) has the same structure and properties than the equivalent rigid arm, any trajectory tracking control method for rigid manipulators can be applied. One of the most widely used techniques for trajectory tracking of robot manipulators is the so-called computed-torque control through feedback linearization (Craig, J.J. 1986) . This method requires an accurate knowledge of the system dynamics. The presence of parametric uncertainties (due to estimation errors or time-varying parameters) or the unmodeled dynamics and disturbances that can affect the plant, can lead to poor or unstable control performance. Therefore, in the present work, a robust controller is applied to the slow subsystem. This controller was proposed for rigid manipulators in (Barambones, O. & Etxebarria, V. 2001). The control law consists of an adaptive feedback linearization term for the modeled dynamics of the system and a robust sliding control to overcome the uncertainties. Concerning the sliding control term, two specific characteristics need to be mentioned. First, the proposed sliding control prevents the chattering effect by smoothing out the control law, substituting the usual sign function by a saturation function. This chattering is especially undesirable for the slow subsystem, since it can result in the excitation of high frequency dynamics. This change introduces a boundary layer around the switching surface. The second particularity consists of an adaptive update of the sliding gain. This avoids the necessity of a prior knowledge of an upper bound of the unmodeled dynamics and noise magnitudes.

The expression of the control law for the slow subsystem is given by:

$$\bar{\tau} = \hat{\bar{B}}_r^{-1}\left[\hat{\bar{M}}_{rr}\ddot{E}_f + \hat{\bar{V}}_{rr}\dot{E}_f + \hat{\bar{G}}_r + \hat{\bar{F}}_r\right] - \hat{P}sat\left(\frac{S}{\beta}\right) = Y(\bar{q}_r,\dot{\bar{q}}_r,\dot{E}_f,\ddot{E}_f)\hat{A} - \hat{P}sat\left(\frac{S}{\beta}\right) \quad (17)$$

where the first term is the estimated nonlinear model-based feedback introduced to compensate for the nonlinearities present in the robot. This term can be expressed as a regressor matrix $Y(\bar{q}_r,\dot{\bar{q}}_r,\dot{E}_f,\ddot{E}_f)$ multiplied by an estimated parameter vector $\hat{A}$. The variable $E_f$ is defined by $\dot{E}_f = \dot{q}_d - \lambda E$ and $E = \bar{q}_r - q_d$ is the tracking error ($q_d$ is the desired trajectory vector). On the other hand, the second term is the sliding control, where $\hat{P}$ is a diagonal matrix formed by the elements of the switching gain vector $\hat{\rho}$ (i.e. $\hat{P} = diag(\hat{\rho})$); $S$ is a surface vector defined by

$$S = \dot{E} + \lambda E = \dot{\bar{q}}_r - \dot{q}_d + \lambda E = \dot{\bar{q}}_r - \dot{E}_f$$

and $\beta = [\beta_1,\ldots,\beta_n]^T, \beta_i > 0$ are the thickness of the boundary layers for each sliding surface associated with each joint. The saturation vector is defined by:

$$sat\left(\frac{S}{\beta}\right) = \left[sat\left(\frac{s_1}{\beta_1}\right),\ldots,sat\left(\frac{s_n}{\beta_n}\right)\right]^T \quad (18)$$

and the saturation function is given by the usual expression:

$$sat\left(\frac{s_i}{\beta_i}\right) = \begin{cases} \text{sgn}(s_i) & \text{if } |s_i| > \beta_i \\ s_i/\beta_i & \text{otherwise} \end{cases} \quad (19)$$

The system dynamical parameters and the elements of the switching gain vector are updated according to the following laws:

$$\dot{\hat{A}} = -\Gamma Y^T(\bar{q}_r,\dot{\bar{q}}_r,\dot{E}_f,\ddot{E}_f)S_0$$
$$\dot{\hat{\rho}} = |S_0|, \quad \hat{\rho}(0) = [0,\ldots,0]^T \quad (20)$$

where $S_0$ is defined by $S_0 = S - Bsat(S/\beta)$ with $B = diag(\beta)$, and $|S_0| = \left[|s_{0_1}|,\ldots,|s_{0_n}|\right]^T$.

It is important to point out that the components of $s_{0_i}$ of $S_0$ are a measure of the distance between the ith component of the surface vector $S$ and the boundary layer (i.e. the interval $[-\beta_i,\beta_i]$), and that outside this boundary layer $\dot{s}_{0_i} = \dot{s}$ is verified. Within the boundary layer, there is not updating of the model parameters and the sliding gains. Moreover, the sliding term turns here into a proportional gain (consequently acting as a PD control law). This behavior is, however, not only not undesirable but very convenient. The adaptation process and the robust control action are only needed if the performance of the closed-loop system is far from the required goal. In the particular case of the slow subsystem it is crucial to select an adequate value of $\beta$ (high enough) in order to avoid the chattering, which can excite high frequency dynamics. Note that the system decomposition into a slow and a fast subsystem has been performed under the time scale separation assumption. If the slow subsystem along with its control shows high frequency dynamics, the singular perturbation approach is no longer valid. Therefore, the control is tuned to take advantage, when needed, of the robustifying properties of the sliding term, but without introducing excessive control activity in the system unnecessarily, so that the time scale separation between the slow and fast dynamics are maintained.

If the measured joint angles are chosen to be the rigid coordinates $q_r$ (so that by Property 3 in section 2 $\bar{B}_r$ is the identity matrix), and assuming that it always can be defined a sufficiently high (unknown) finite nonnegative gain vector $\rho = [\rho_1,\ldots,\rho_n]^T$ such that $\rho \geq \bar{V}_{rr}\beta_{\max} + \eta$, where $\eta$ is a vector of positive elements, then the following stability result can be formulated:

**Theorem:** The control law (17) with the adaptation mechanisms (20) lead the rigid variables $\bar{q}_r$ of the slow system (12) and their derivatives $\dot{\bar{q}}_r$ asymptotically towards the desired trajectories $q_d$ and their derivatives $\dot{q}_d$. Moreover, the tracking error vector $E = \bar{q}_r - q_d$ can be made as small as desired by choosing adequately small boundary layers $\beta_i$.

**Proof:** Define the following Lyapunov function candidate:

$$V = \frac{1}{2}\left[S_0^T\bar{M}_{rr}S_0 + \tilde{A}^T\Gamma^{-1}\tilde{A} + \tilde{\rho}^T\tilde{\rho}\right] \quad (21)$$

where the parametric errors $\tilde{A} = \hat{A} - A$ and $\tilde{\rho} = \hat{\rho} - \rho$ have been defined. Since we are outside the boundary layer, where $\dot{S}_0 = \dot{S}$, it follows that:

$$\dot{V} = S_0^T \bar{M}_{rr} \dot{S}_0 + \frac{1}{2} S_0^T \dot{\bar{M}}_{rr} S_0 + \tilde{A}^T \Gamma^{-1} \dot{\tilde{A}} + \tilde{\rho}^T \dot{\tilde{\rho}} =$$

$$= S_0^T \bar{M}_{rr} \dot{S} + \frac{1}{2} S_0^T \dot{\bar{M}}_{rr} S_0 + \tilde{A}^T \Gamma^{-1} \dot{\tilde{A}} + \tilde{\rho}^T \dot{\tilde{\rho}} =$$

$$= S_0^T \bar{M}_{rr} \ddot{q}_r - S_0^T \bar{M}_{rr} \ddot{E}_f +$$

$$+ \frac{1}{2} S_0^T \dot{\bar{M}}_{rr} S_0 + \tilde{A}^T \Gamma^{-1} \dot{\tilde{A}} + \tilde{\rho}^T \dot{\tilde{\rho}}$$

Using (12) $\bar{M}_{rr} \ddot{q}_r$ can be rewritten to yield:

$$\dot{V} = S_0^T \left[ \bar{\tau} - \bar{V}_{rr} \dot{q}_r - \bar{F}_r - \bar{G}_r - \bar{M}_{rr} \ddot{E}_f \right] +$$

$$+ \frac{1}{2} S_0^T \dot{\bar{M}}_{rr} S_0 + \tilde{A}^T \Gamma^{-1} \dot{\tilde{A}} + \tilde{\rho}^T \dot{\tilde{\rho}} =$$

$$= S_0^T \left[ \bar{\tau} - \bar{V}_{rr}(S + \dot{E}_f) - \bar{F}_r - \bar{G}_r - \bar{M}_{rr} \ddot{E}_f \right] +$$

$$+ \frac{1}{2} S_0^T \dot{\bar{M}}_{rr} S_0 + \tilde{A}^T \Gamma^{-1} \dot{\tilde{A}} + \tilde{\rho}^T \dot{\tilde{\rho}} =$$

$$= S_0^T \left[ \bar{\tau} - \bar{V}_{rr}(S_0 + Bsat(S/\beta) + \dot{E}_f) - \bar{F}_r -$$

$$- \bar{G}_r - \bar{M}_{rr} \ddot{E}_f \right] + \frac{1}{2} S_0^T \dot{\bar{M}}_{rr} S_0 + \tilde{A}^T \Gamma^{-1} \dot{\tilde{A}} + \tilde{\rho}^T \dot{\tilde{\rho}} =$$

$$= S_0^T \left[ \bar{\tau} - \bar{M}_{rr} \ddot{E}_f - \bar{V}_{rr} \dot{E}_f - \bar{F}_r - \bar{G}_r - \bar{V}_{rr} Bsat(S/\beta) \right] +$$

$$+ \frac{1}{2} S_0^T \left[ \dot{\bar{M}}_{rr} - 2\bar{V}_{rr} \right] S_0 + \tilde{A}^T \Gamma^{-1} \dot{\tilde{A}} + \tilde{\rho}^T \dot{\tilde{\rho}}$$

Now, since $\left[ \dot{\bar{M}}_{rr} - 2\bar{V}_{rr} \right]$ is antisymmetric by Property 2 of section 2, its associated quadratic form is zero. Moreover, expressing the equation in terms of the regressor $Y$ one obtains:

$$\dot{V} = S_0^T \left[ \bar{\tau} - YA - \bar{V}_{rr} Bsat(S/\beta) + \tilde{A}^T \Gamma^{-1} \dot{\tilde{A}} + \tilde{\rho}^T \dot{\tilde{\rho}} \right]$$

And by the control law (17), taking into account we are outside the boundary layer:

$$\dot{V} = S_0^T \left[ Y\hat{A} - \hat{P} \text{sgn}(S) - YA - \bar{V}_{rr} Bsat(S/\beta) \right] +$$

$$+ \tilde{A}^T \Gamma^{-1} \dot{\tilde{A}} + \tilde{\rho}^T \dot{\tilde{\rho}} = S_0^T Y\tilde{A} - \hat{\rho} |S_0| -$$

$$- S_0^T \bar{V}_{rr} Bsat(S/\beta) + \tilde{A}^T \Gamma^{-1} \dot{\tilde{A}} + \tilde{\rho}^T \dot{\tilde{\rho}}$$

And using the expressions for the adaptive laws (20):

$$\dot{V} = S_0^T Y\tilde{A} - \hat{\rho} |S_0| - S_0^T \bar{V}_{rr} Bsat(S/\beta) -$$

$$- \tilde{A}^T Y^T S_0 + (\hat{\rho}^T - \rho^T) |S_0| = \quad (22)$$

$$= -S_0^T \bar{V}_{rr} Bsat(S/\beta) - \rho^T |S_0| \leq -\eta^T |S_0|$$

Thus $V$ is a positive definite and decrescent function, which implies that the states $S_0$, $\tilde{A}$ and $\tilde{\rho}$ are bounded.

Since $A$ and $\rho$ are bounded, then $q_r$, $\dot{q}_r$, $\hat{A}$ and $\hat{\rho}$ are bounded. Also, since the closed loop dynamics can be written in terms of $S$:

$$\bar{M}_{rr} \dot{S} + \bar{V}_{rr} S = Y\tilde{A} - \hat{P} sat(S/\beta)$$

and as $Y$, $\tilde{A}$, $\bar{V}_{rr}$, $\hat{\rho}$ and $S$ are bounded and $\bar{M}_{rr}^{-1}$ exists by Property 1 of section 2, then $\dot{S}$ is bounded. Also, since $\dot{S}_0 = \dot{S}$ outside the boundary layer (and $\dot{S}_0 = 0$ inside it) then $\dot{S}_0$ is also bounded. Thus from (22) one concludes that $\ddot{V}$ is bounded, so $\dot{V}$ is a uniformly continuous function, and from Barbalat's lemma we can conclude that $\dot{V} \to 0$ as $t \to \infty$, which from (22) implies that $S_0 \to 0$ as $t \to \infty$, or equivalently $S$ converges to the boundary layer asymptotically. Thus, under the definition of $S$, the tracking error of each joint converge to a small size depending on the thickness $\beta_i$ for each joint.

### 3.2 LQR control for the fast system

The fast controller must stabilize the deflections of the flexible modes around the equilibrium manifold for the fast system (16), whose matrices $A_F$ and $B_F$ must form an stabilizable pair for any value of the rigid variables $\bar{q}_r$. This assumption is usually true in flexible manipulators, where the fast dynamics is normally marginally stable. The fast control law has the form:

$$\tilde{\tau} = K_{pf}(\bar{q}_r)\varphi 1 + K_{df}(\bar{q}_r)\varphi 2 \quad (23)$$

where the gains $K_{pf}$ and $K_{df}$ are designed using and LQR strategy in this paper. This optimal approach to the problem is well suited to deal with the damping of the flexible dynamics (16), since the control objective can be conveniently expressed as minimizing a cost function:

$$J = \int \left( \varphi^T Q \varphi + \tilde{\tau}^T R \tilde{\tau} \right) dt \quad (24)$$

where $Q$ and $R$ are the standard LQR weighting matrices, $\varphi = [\varphi 1, \varphi 2]^T$ is the state of the fast subsystem (14), and $\tilde{\tau}$ is the fast part of the control law (10).

In this way, according to Tikhonov's Theorem, once stabilized the fast subsystem, the trajectories of the global system verify:

$$q_r = \bar{q}_r + O(\varepsilon); \quad q_f = \varepsilon^2 (\bar{\psi} + \varphi 1) + O(\varepsilon) \quad (25)$$

so the proposed fast control law, which minimizes $\varphi$ (and thus $\varphi 2$), stabilizes the deflections around the slow manifold.

The choose of the weighting matrices is related to the fast control law amplitude and the corresponding damping of the flexible modes. A suitable equilibrium between both values should be found. On the one hand, increasing the flexible modes damping needs a higher control effort. On the other hand this would result in a higher influence of $\tilde{\tau}$ on the rigid variable evolution $q_r(t)$, disturbing the desired trajectory tracking. These opposed effects should be balanced so that the flexible modes are effectively damped, but without compromising the rigid control performance, in such a way that the time-scale separation between the fast and slow subsystems is maintained in the closed-loop.

## 4. Experimental results

The effectiveness of the proposed control scheme has been tested by means of real time experiments on a laboratory single flexible link. This manipulator arm, fabricated by Quanser Consulting Inc. (Ontario, Canada), is a spring steel bar that moves in the horizontal plane due to the action of a DC motor.

| Property | Value |
|---|---|
| Motor inertia, $I_h$ | 0.002 $Kgm^2$ |
| Link length, $L$ | 0.45 $m$ |
| Link height, $h$ | 0.02 $m$ |
| Link thickness, $d$ | 0.0008 $m$ |
| Link mass, $M_b$ | 0.06 $Kg$ |
| Linear density, $\rho$ | 0.1333 $Kg/m$ |
| Flexural rigidity, $EI$ | 0.1621 $Nm^2$ |

Table 1. Flexible link parameters

A potentiometer measures the angular position of the system, and the arm deflections are measured by means of a strain gauge mounted near its base. The whole system, whose parameters are shown in Table 1, is displayed in Fig. 1.

The dynamical modelling of the arm has been carried out through the measurement of the natural mode shapes and frequencies. The resulting state-space representation capturing the first two flexible modes is:

$$\dot{x}(t) = \begin{bmatrix} 0 & 0 & 0 & 1 & 0 & 0 \\ 0 & 0 & 0 & 0 & 1 & 0 \\ 0 & 0 & 0 & 0 & 0 & 1 \\ 0 & 0 & 0 & 0 & 0 & 0 \\ 0 & -\omega_1^2 & 0 & 0 & -2\delta\omega_1 & 0 \\ 0 & 0 & -\omega_2^2 & 0 & 0 & -2\delta\omega_2 \end{bmatrix} x(t)$$
$$+ \begin{bmatrix} 0 & 0 & 0 & \dfrac{1}{I_t} & \phi_1'(0) & \phi_2'(0) \end{bmatrix}^T \tau$$

where $\phi_1(x)$ and $\phi_2(x)$ are the first two mode shapes whose respective frequecies are $\omega_1 = 21.80$ rad/s and $\omega_2 = 128.80$ rad/s. Also, $\phi_1'(0)$ and $\phi_2'(0)$ are the first spatial derivatives of $\phi_1(x)$ and $\phi_2(x)$, respectively, evaluated at the base of the robot. It should be noted here that although this model is very ideal, and will not reflect all the complexities of the real experimental system, it is enough to design satisfactory controllers, as it will be shown by the experimental results.

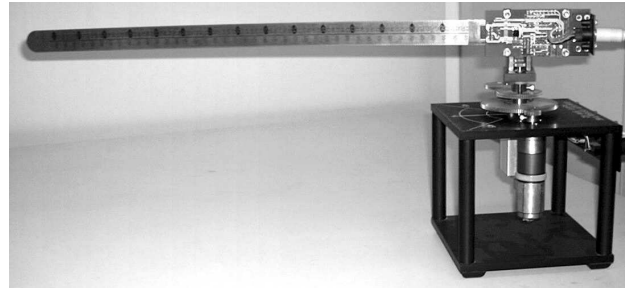

Fig. 1. Photograph of the experimental flexible arm

In order to design the composite control law, the arm model is decomposed into the slow and the fast subsystem. The singular perturbation parameter $\varepsilon$ is the inverse of $\omega_1$, that is $\varepsilon = 0.0459$. For the design of the slow controller the friction effects acting on the rigid variables have been considered as unmodeled dynamics, to test the robustness properties of the sliding-mode slow controller. Finally the complete slow control has been implemented as shown in section 3.1 with $\lambda = 10$ and $\beta = 1.3$. Regarding the fast LQR control, the design has been carried out solving the Riccati equations using the Control System Toolbox from Matlab (Mathworks, Inc., 2002) with the weighting matrices $Q = diag(150,500,1,0)$ and $R = 2$.

Experimental results are shown on Figs. 2 and 3. In Fig. 2 the control results using a rigid (slow-only) control design are displayed. As shown, the rigid variable tracks the reference (with a certain error), but the naturally excited flexible vibrations are not well damped (Fig. 2(c)), which is understandable as the manipulator is being treated here as if it were rigid. Also, within this rigid-only control framework, a stronger control action (in an attempt to make the rigid tracking error smaller) would, in turn, significantly worsen the flexible part response (the flexible vibrations would then be larger). Note that this rigid-only control results are similar to those obtained by conventional robot controllers such as PIDs.

Fig. 3 shows the experimental results obtained when using the proposed combined rigid-flexible (slow-fast) control strategy. As seen in the graphics, the rigid variable accurately follows the desired trajectory, and moreover the flexible modes are now conveniently damped (compare Fig. 2(c) and Fig. 3(c)). This results in negligible deflections during the steady intervals of the rigid trajectory (i.e.: as shown in Fig. 3(c), vibrations are close to zero in the intervals which go from 1 to 2 seconds, 3 to 4 s., 5 to 6 s., 7 to 8 s., etc., which coincide with the steady up and down positions of the rigid variable). Also, the rigid part of the control can be retuned to get smaller rigid tracking errors (compare

Figs. 2(a) (b) and Figs. 3(a) (b)) without compromising the flexible deflections damping. This reinforces the conclusion that the proposed combined sliding-LQR design provides better tracking properties than conventional robot control schemes, both in the rigid and in the flexible responses.

## 5. Conclusions

In this paper a composite robust-optimal controller for flexible-link manipulators has been proposed. As a first step, the system dynamics has been approximated by two reduced order subsystems by means of the singular perturbation theory. In order to take into account the uncertainties, unmodeled dynamics and disturbances that can affect the system, the controller has been designed using robust techniques. The slow subsystem, which describes the equivalent rigid arm dynamics, has been controlled by means of a sliding-mode controller. The dynamic knowledge is introduced in the control law through a feedback linearization term, but the model parameters are updated based only on the measured performance. And additional sliding term in the control law deals with the uncertainties and noise.

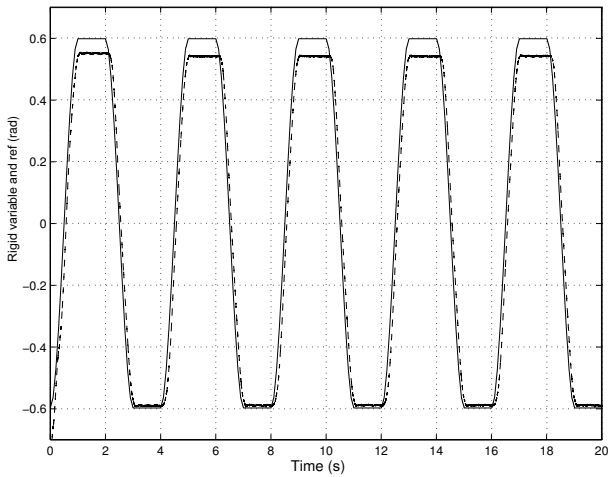

(a)

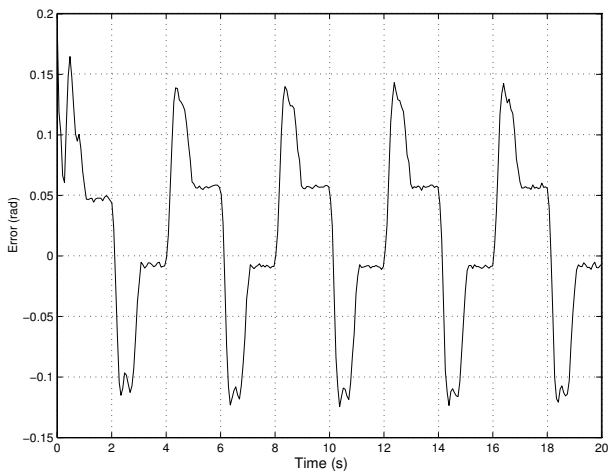

(b)

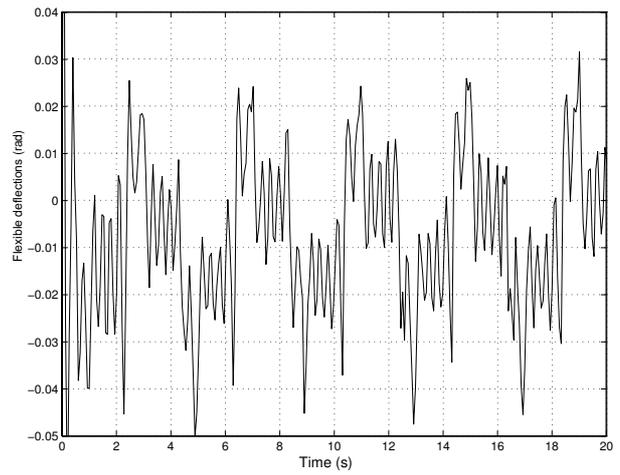

(c)

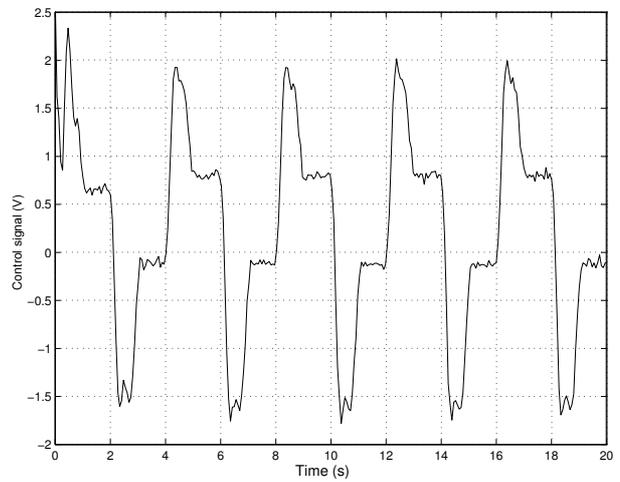

(d)

Fig. 2. Experimental results for pure rigid control: (a) Time evolution of the rigid variable $q_r$ -- and reference $q_d$ -; (b) Rigid variable tracking error; (c) Time evolution of the flexible deflections; (d) Control signal

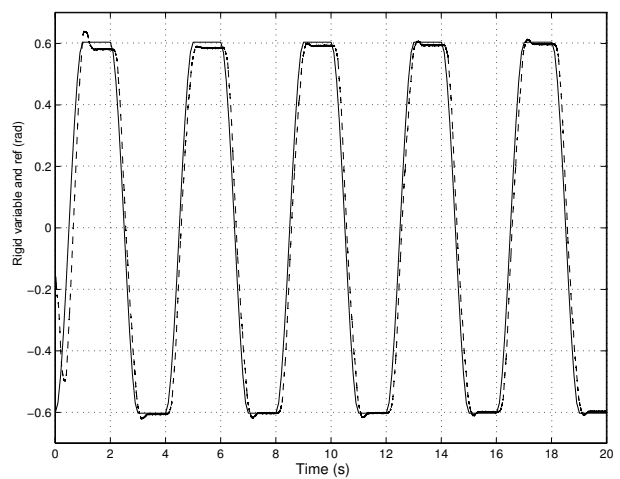

(a)

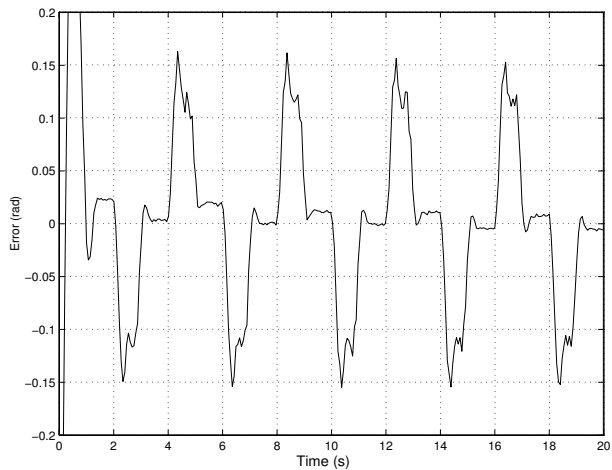

(b)

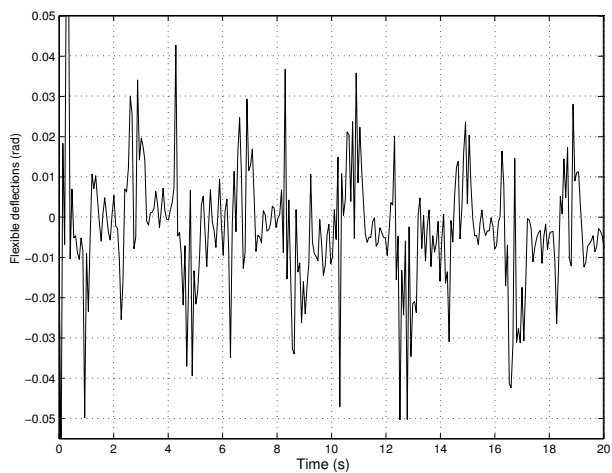

(c)

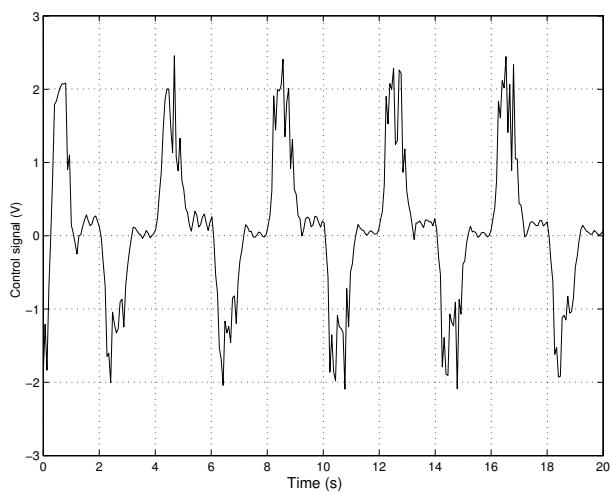

(d)

Fig. 3. Experimental results for composite (slow-fast) sliding-LQR control: (a) Time evolution of the rigid variable $q_r$ -- and reference $q_d$ -; (b) Rigid variable tracking error; (c) Time evolution of the flexible deflections; (d) Composite control signal

For the fast subsystem an LQR control strategy has been proposed. This design takes into account the coupling between the slow and fast dynamics by means of a disturbance term affecting the input of the fast subsystem. The complete controller has been tested by means of real time experiments on a laboratory flexible arm. The experimental results have illustrated the suitability of the proposed control scheme, which has been shown to hold superior tracking properties and adaptation capabilities with respect to conventional rigid robotic control designs, while simultaneously damping conveniently the naturally excited flexible vibrations.

**Acknowledgment:** The authors are grateful to UPV/EHU for partial support of this work through projects UPV00224.310-E-14877 and 9/UPV 00224.310-15254/2003. They are also grateful to the Basque Government for grant BFI04.440.